\documentclass{article}

    \PassOptionsToPackage{numbers, compress}{natbib}

\usepackage[preprint]{neurips_2024}

\usepackage[utf8]{inputenc} 
\usepackage[T1]{fontenc}    
\usepackage{hyperref}       
\usepackage{url}            
\usepackage{booktabs}       
\usepackage{amsfonts}       
\usepackage{nicefrac}       
\usepackage{microtype}      
\usepackage{xcolor}         
\usepackage{amsmath} 
\usepackage{graphicx}
\usepackage{caption}
\usepackage{wrapfig,lipsum}

\title{Llama 3 Meets MoE: Efficient Upcycling}

\author{%
  Aditya Vavre$^{1}$\thanks{Work done as part of internship at NVIDIA}
  \quad
  Ethan He$^2$\thanks{Corresponding author: yihuih@nvidia.com}
  \quad
  Dennis Liu$^2$ 
  \quad
  Zijie Yan$^2$
  \quad
  \textbf{June Yang}$^2$ \\
  \quad
  \textbf{Nima Tajbakhsh}$^2$
  \quad
  \textbf{Ashwath Aithal}$^2$ \\
  $^1$Univeristy of Texas at Austin \quad $^2$NVIDIA
}

\begin{document}

\maketitle

\begin{abstract}
   Scaling large language models (LLMs) significantly improves performance but comes with prohibitive computational costs. Mixture-of-Experts (MoE) models offer an efficient alternative, increasing capacity without a proportional rise in compute requirements. However, training MoE models from scratch poses challenges like overfitting and routing instability. We present an efficient training recipe leveraging pre-trained dense checkpoints, training an 8-Expert Top-2 MoE model from Llama 3-8B with less than 1\% of typical pre-training compute. Our approach enhances downstream performance on academic benchmarks, achieving a \textbf{2\%} improvement in 0-shot accuracy on MMLU, while reaching a Model FLOPs Utilization (MFU) of \textbf{46.8\%} during training using our framework. We also integrate online upcycling in NeMo\footnote{https://github.com/NVIDIA/NeMo} for seamless use of pre-trained weights, enabling cost-effective development of high-capacity MoE models.
\end{abstract}

\section{Introduction}
Transformers~\cite{attention} have rapidly become the foundational architecture for a wide range of tasks in natural language processing~\cite{fewshotlearners, bert, T5, gpt} and computer vision~\cite{vit, registers, scalingvision}, revolutionizing these fields with their scalability and remarkable effectiveness. This has driven a dramatic increase in model complexity, with modern implementations featuring billions of parameters, far exceeding earlier architectures~\cite{palm, T5, megatronlm}. Such growth has been underpinned by established scaling laws, which show that the cross-entropy loss follows a power-law relationship with model size, dataset size, and compute resources allocated for training~\cite{scalinglaws}. This relationship allows us to determine the optimal allocation of a fixed compute budget. Despite these advantages, scaling large language models (LLMs) to billions or trillions of parameters is not without challenges. For instance, GPT-4 required ~55M GPU hours to train the model on 25,000 A100 GPUs amounting to more than 50M dollars in training expenditure. These challenges have fueled interest in approaches like Mixture-of-Experts (MoE) architectures, which aim to enhance model capacity without proportionally increasing computational costs~\cite{sparsemoe, switch-transformer}. However, training MoE models from scratch remains a complex endeavor. It can be time consuming and costly with potential issues such as over-fitting, routing instability and expert collapse~\cite{mixtral, sparsemoe}. Moreover, MoE models often need a larger and more diverse dataset for effective training. By providing an efficient recipe and enabling the use of pre-trained checkpoints to initialize MoE models, developers can achieve higher-performing models with relatively modest compute budgets. 

Our contribution is summarized as follows: 

\begin{enumerate}
    \item We train a 8-Expert Top-2 (E8T2) MoE model starting from the Llama 3-8B model on an academic dataset blend. We propose a training framework and release a recipe for efficiently training MoE models with <1\% of the pre-training compute.
    \item We show improvement in downstream task performance on commonsense reasoning and knowledge benchmarks such as MMLU. 
    \item We conducted two ablation experiments to validate our selection of the capacity factor and routing algorithm for training.
    \item We implement online upcycling in NeMo allowing for the use of pre-trained model weights to initialize and train MoE models. 
\end{enumerate}

\section{Background}
\begin{table}[t!]
\centering
\small
\caption{Comparison of total model parameters, activated model parameters and FLOPs during a forward pass of a base Llama 3-8B and Llama 3-E8T2 model. BS refers to batch size.}
\begin{tabular}{lccc}
\toprule
\textbf{Model} & \textbf{Total params} & \textbf{Active params} & \textbf{FLOPs (BS=1)} \\ \midrule \midrule
Llama 3-8B & 8B & 8B & 4.7e14 \\ \midrule
Llama 3-E8T2 & 34.4B & 11.8B & 7.5e14 \\ \bottomrule
\end{tabular}
\label{tab:activated_params}
\end{table}
Mixture of Experts (MoE) is an ensemble learning technique that scales model capacity without significantly increasing training or inference costs. In MoE models, the MLP layers in a transformer block are typically replaced with several “experts” $E_1, \ldots E_N$ that have distinct learnable parameters. A small gating network $G$ called the “router” controls which set of experts receive a particular token.

Let $G(x)$ and $E_i(x)$ denote the output of the router and the $i^{th}$ expert on an input $x$ respectively. The output $y$ of the MoE is given by:
\begin{equation}
    y = \sum_{i=1}^{N} G(x)_iE_i(x)
\end{equation}
Several routing algorithms have been developed for example, Top-k~\cite{sparsemoe} and Expert Choice~\cite{expertchoicerouting}. In the sparse setting, only a subset of experts ($k$) is activated which is much smaller than the number of total experts ($N$) to save compute~\cite{sparsemoe}. This is done through a Noisy Top-k Gating as shown below:

\begin{equation}
    G(x) = Softmax(KeepTopK(H(x), k)) 
\end{equation}
\begin{equation}
    H(x)_i = (x\cdot W_g)_i + StandardNormal() \cdot Softplus((x\cdot W_{noise})_i)
\end{equation}
\begin{equation}
    KeepTopK(v,k)_i = 
    \begin{cases}
    v_i,              & \text{if $v_i$ is in the top $k$ elements of $v$} \\
    -\infty,              & \text{otherwise}
    \end{cases}
\end{equation}
    
$W_g$ denotes the weight matrix of the router. The amount of noise per component is controlled by a second trainable weight matrix $W_{noise}$, which helps with load balancing~\cite{sparsemoe}. 
To further ensure efficient training, the average token load per expert is regulated by a capacity factor (CF)~\cite{switch-transformer}. 
\begin{equation*}
    \text{expert capacity} = \frac{\text{tokens per batch}}{\text{$N$}}\times \text{CF}
\end{equation*}
Overflowing tokens assigned to experts are excluded from computation and directly routed to the layer's output. The CF controls trade-off between the performance of the MoE model and its accuracy. Increasing the CF increases the quality but increases communication costs and memory of activations~\cite{switch-transformer}. To further optimize MoE training, expert parallelism (EP) is employed. In EP, different experts are placed on different devices, and executed in parallel. This allows us to increase the number of experts (and hence the number of parameters) by
proportionally increasing the number of devices in the training cluster~\cite{sparsemoe}. 

MoEs can dramatically increase the number of parameters without a proportional increase in computational cost. \autoref{tab:activated_params} shows that an 8-Expert Top-2 (E8T2) MoE model, despite being approximately 4$\times$ larger in size, utilizes only about 1.6$\times$ the total FLOPs of its non-MoE counterpart during a forward pass. However, training MoE models from scratch can still pose a problem due to data hungry and instability issues during training~\cite{sparseupcycling}. Sparse Upcycling\cite{sparseupcycling} is a way to reuse the
sunk training costs of a dense language model by initializing a sparsely activated MoE model from a dense checkpoint.

\section{Method}
\label{upcycle_algo}
\subsection{Upcycling Technique}
\begin{figure}
\centering
    \includegraphics[width=1.0\linewidth]{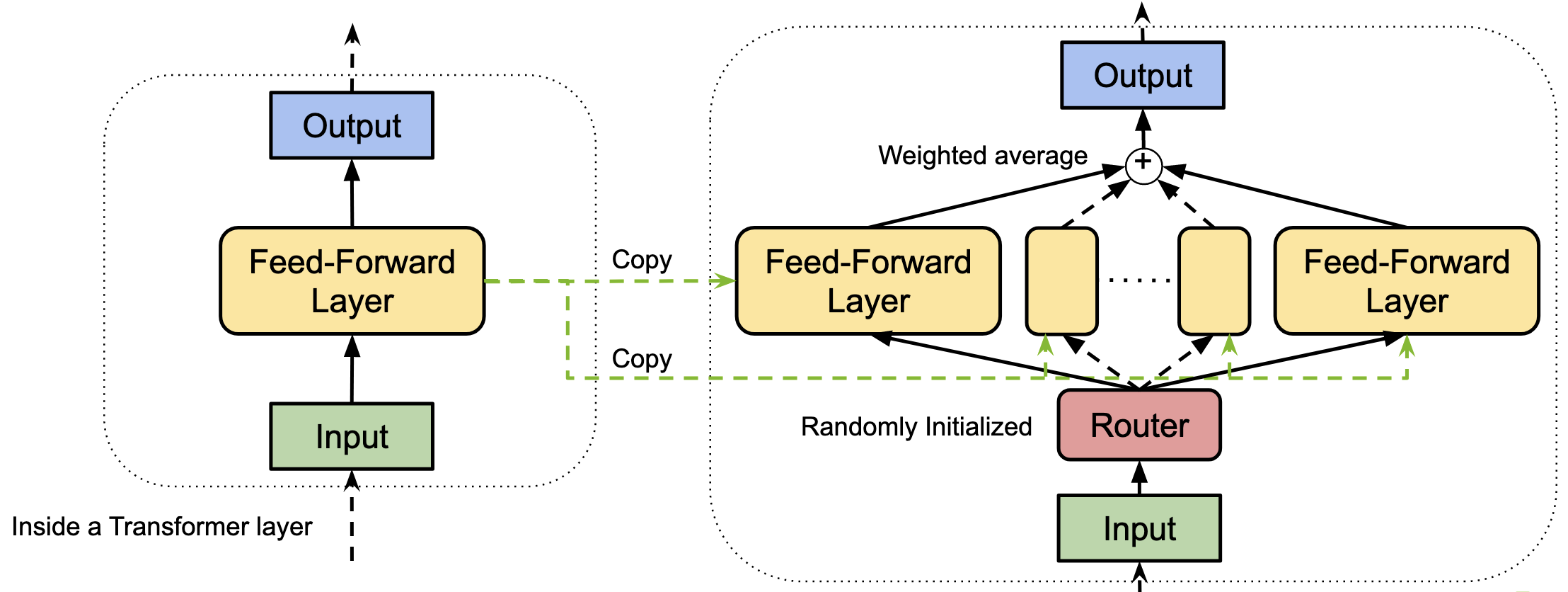}
    \caption{Our upcycling method. The feedforward layer is copied over $N$ times to initialize the experts in the MoE model and the router is randomly initialized.}\looseness=-1
    \label{fig:method_main}
\end{figure}

The upcycling method is illustrated in \autoref{fig:method_main}. We follow a similar approach to \cite{sparseupcycling} and \cite{upcyclinglargelanguagemodels}. We assume we have access to a dense checkpoint of a pre-trained language model. We convert a subset of the feed-forward layers in the dense model to MoE layers. To upcycle a feed-forward layer to a $N$ Expert Top-$k$ MoE layer, we simply copy the weights of the feed-forward layer $N$ times to initialize the experts i.e., each expert is a copy of the original feed-forward layer. Additionally, we add a router which is initialized with random weights. All of the remaining weights including the embedding layer are simply copied from the dense checkpoint. 

Upcycling can be challenging to implement in a distributed training setting when the dense checkpoint contains a large number of parameters, as is typical with large language models (LLMs). This is because upcycling significantly increases the total parameter count, and each device typically holds a full copy of the shared model parameters and gradients, which can cause memory requirements to exceed the capacity of each node. To address this issue, we implement online upcycling in NeMo, enabling users to upcycle by supplying a dense checkpoint and a parallel training configuration. For efficient implementation, the dense checkpoint is sharded based on the specified parallel training configuration, and weights are upcycled independently on each device, avoiding additional computation and eliminating the need for cross-device weight copying. We make our implementation available in NeMo\footnote{https://github.com/NVIDIA/NeMo}.  

\subsection{Training Framework}
To efficiently train the MoE models at scale, we leveraged 5-D hybrid parallelisms with Megatron-Core\footnote{https://github.com/NVIDIA/Megatron-LM}, which supports Tensor Parallelism(TP;~\cite{Megatron-TP, Megatron-SP}), Expert Parallelism(EP;~\cite{switch-transformer, gshard}), Pipeline Parallelism(PP;~\cite{gpipe}), Context Parallelism(CP;~\cite{ring-attention}) and Data Parallelism(DP;~\cite{zero-dp, fsdp}) for distributed training with thousands of GPUs. 
TP shards the tensors of each layer to difference devices. EP splits experts in the MoE layer into multiple devices. PP cuts the transformer layers into multiple stages.
CP separates the input sequence into multiple segments, reducing the memory footprint for long-sequence training. And we use DP with ZeRO-1 to scales model training further by replicates model weights and shard optimizers states across DP ranks.

To further improve training efficiency, we introduced MoE Parallel Folding, a heterogeneous hybrid parallel strategy that decouples the Attention and MoE components of the Transformer. The core idea is to decouple the parallel mapping of the Attention and MoE layers to potentially enhance performance. For the Attention layer, we establish 4-dimensional parallel mappings comprising TPxCPxDPxPP. For the MoE layer, we create another 4-dimensional parallel group consisting of Expert-TPxEPxExpert-DPxPP. This allows for setting arbitrary and separate TP, EP, CP, DP sizes for the MoE and Attention components. With MoE Parallel Folding, the communication-intensive parallelism from the Attention and MoE layer can be folded together and fit into the high-bandwidth NVLink domain as much as possible, which greatly reduces the communication overhead. For example, we can set TP2CP2 for the attention layer and TP1EP8 for the MoE layer; then the TP and CP group the in attention layer can be folded into the EP group within a single node with 8 GPUs.

\begin{table}[t!]
    \caption{Training performance for different configurations. CF indicates the capacity factor for each expert. The last row indicates training without dropping tokens.}
    \label{tab:performance-metrics}
    \centering
    \begin{tabular}{ccccccccccccc}
        \toprule
        \textbf{GPUs} & \textbf{CF} & \textbf{TP} & \textbf{CP} & \textbf{Expert-TP} & \textbf{EP} & \textbf{PP} & \textbf{VP} & \textbf{TFLOPS/GPU} & \textbf{MFU} \\
        \midrule
        128 & 1 & 1 & 2 & 1 & 8 & 4 & 8 & 462.8 & 46.8\% \\
        128 & 2 & 2 & 2 & 1 & 8 & 4 & 8 & 387.5 & 39.2\% \\
        128 & 4 & 2 & 2 & 1 & 8 & 4 & 8 &  389.7 & 39.4\% \\
        128 & N/A & 2 & 2 & 1 & 8 & 4 & 8 & 391.8 & 39.6\% \\
        \bottomrule
    \end{tabular}
\end{table}

\textbf{}
By tuning of parallelism mappings, we can achieve 39.2-46.8\% Model FLOPs Utilization(MFU) for different configurations in ~\autoref{tab:performance-metrics}. Training with capacity factor 1 has better MFU than dropless training, since they can prevent the load imbalance issue and have less memory footprint to enable smaller model parallelism. There are some tuning practices to find the best configurations for MoE model training:
\begin{enumerate}
\item TP and EP involve significant communication overhead at each layer, making it advantageous to keep them within the NVLink domain to minimize latency. For MoE layers specifically, EP generally outperforms TP in terms of performance.
\item In Megatron-Core, there are two types of token dispatchers; the AllGather-based token dispatcher and the AllToAll-based token dispatcher. Usually, the latter is more efficient for MoE models with smaller routing TopK values, such as 1-4.
\item In long-context LLM training, CP can be utilized to reduce memory usage and improve efficiency by overlapping communication and computation. This approach is particularly effective with Grouped-Query Attention (GQA), which reduces communication overhead due to the smaller message size of KV features.
\item Scaling across nodes with PP and DP is advantageous. Introducing Virtual Pipeline Parallelism (VPP) further enhances performance by reducing the pipeline bubble size.
\item Enabling recomputation for MoE layers during the early training stage helps mitigate out-of-memory issues caused by severe load imbalances.
\end{enumerate}

\section{Experiments}

\subsection{Training Dataset}
The training data for the upcycling experiments consists of two sources. The first is the RedPajama V2 pretraining data which is deduplicated and filtered. We then divide the data into 3 buckets based on the n-gram perplexity following CCNet pipeline~\cite{ccnet}. We only use the bucket with the least perplexity for training, which contains about 0.89T tokens. The second training data source is academic data, a blend of various open-source academic benchmark datasets, comprising approximately 2.7 billion tokens~\cite{nemotron}. We use a blend of two sources in a 7:3 ratio. 

\subsection{Experimental Setup}
In  this section we describe the experimental setup. We begin all upcycling experiments from the Llama 3-8B pretrained checkpoint. We upcycle Llama 3-8B model weights to create an 8-Expert Top-2 (E8T2) MoE model using the technique described in \autoref{upcycle_algo}. We use a CF of 4, 8-way expert parallelism, 2-way tensor parallelism, 4-way pipeline parallelism, 8-way virtual pipeline parallelism and data parallelism to train the model. We train the model on 100B tokens starting from a learning rate of 3e-5 decayed to 3e-7 using a cosine annealing scheduler with 100 warmup steps. The training was performed on 512 H100 GPUs using 16-bit floating point (\texttt{bfloat 16}) precision.

\section{Results and Analysis}
In this section we describe the results of Llama 3-8B base and our upcycled Llama 3-8B E8T2 model on some common academic benchmarks. In line with several prior works, we use lm-evaluation-harness to evaluate and report normalized accuracy on the following tasks: 5-shot and 0-shot MMLU~\cite{mmlu}, 0-shot TruthfulQA~\cite{truthfulqa}, PIQA~\cite{piqa}, SciQ~\cite{sciq}, LogiQA~\cite{logiqa}, BoolQ~\cite{boolq} and OpenBookQA~\cite{openbookqa}.
The results are summarized in ~\autoref{tab:main_results}. We see an improvement of 2\% on MMLU 0-shot score and $\sim$1.2\% overall improvement over the base Llama 3-8B model.

\begin{table*}[t!]
\centering
\small
\addtolength{\tabcolsep}{-3.0pt}
\caption{Normalized accuracy of Llama 3-8B Base Model vs upcycled Llama 3-8B 8 Expert Top-2 MoE model on downstream tasks. All reported numbers are 0-shot performance unless specified in brackets.}
\begin{tabular}{lcccccccccc}
\toprule
\textbf{Model} & \textbf{MMLU(5)} & \textbf{MMLU} & \textbf{TruthfulQA} & \textbf{PIQA} & \textbf{SciQ} & \textbf{LogiQA} & \textbf{BoolQ} & \textbf{OBQA} & \textbf{Average} \\ \midrule \midrule
Llama 3-8B & 65.20 & 62.10 & 44.01 & 80.47 & 93.90 & 29.80 & 81.16 & 45.00 & 62.71 \\ \midrule
Llama 3-8B E8T2 & 64.00 & 64.10 & 44.22 & 78.62 & 97.00 & 30.11 & 88.23 & 44.80 & 63.89 \\ \bottomrule
\end{tabular}
\label{tab:main_results}
\end{table*}

Notably, our upcycling process on 100B tokens consumed ~11K GPU hours, compared to an estimated 1.6 million GPU hours required to train the MoE model from scratch on the entire Llama 3 training dataset on 512 H100 GPUs. Upcycling massively reduces the training costs by recycling the previously invested GPU hours, enabling users to obtain models with better downstream performance. In the following sections, we discuss our choice of capacity factor and the routing algorithm used during training.  

\subsection{Choice of Capacity Factor}
\begin{table}[t]
\centering
\small
\caption{Model Flops Utilization (MFU) and MMLU accuracy of Llama 3-8B base model continued training (CT) vs upcycled Llama 3-8B E8T2 model with different capacity factors (CF). Dropless refers to an infinite CF. }
\begin{tabular}{lccc}
\toprule
\textbf{Training Strategy} & \textbf{MFU(\%)} & \textbf{MMLU(5)} & \textbf{MMLU} \\ \midrule \midrule
Base Model CT & 52.4 & 62.4 & 62.9 \\ \midrule
Dropless & 39.6 & 63.3 & 63.7 \\ \midrule 
CF 4 & 39.4 & 63.5 & 63.8 \\ \midrule 
CF 2 & 39.2 & 64.0 & 63.9 \\ \midrule 
CF 1 & 46.8 & 63.7 & 63.3 \\ \bottomrule
\end{tabular}
\label{tab:cf_ablation}
\end{table}
\begin{figure}
\centering
    \includegraphics[width=0.6\linewidth]{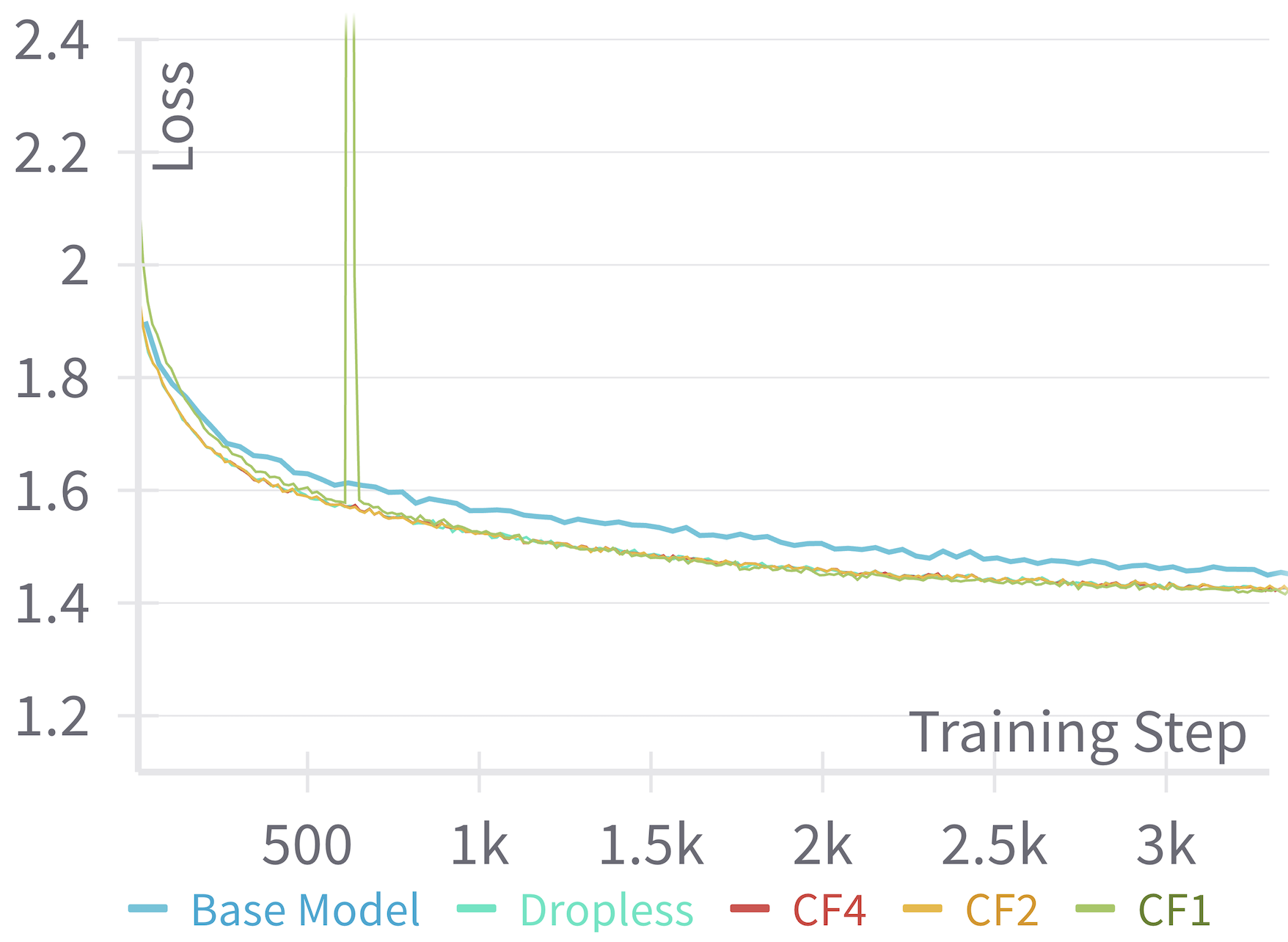}
    \caption{Training loss of Llama 3-8B base model continued training (CT) vs upcycled Llama 3-8B E8T2 model with different capacity factors (CF).}\looseness=-1
    \label{fig:cf_ablation_loss}
\end{figure}

We compare the trade-off between performance and accuracy using different capacity factors to train the MoE model. ~\autoref{tab:cf_ablation} compares the performance measured in model flops utilization (MFU) and accuracy on MMLU for varying capacity factors. We train using the same blend of data for 27B tokens. The loss curve is shown in~\autoref{fig:cf_ablation_loss}. We compare different CF settings against the base model CT in which the base Llama 3-8B model is trained without upcycling to a MoE model. We also compare the MFU and accuracy of a MoE model trained without a capacity factor, using a "token-dropless" approach where no overflowing tokens are dropped. As expected, the base model CT has the highest MFU followed by training with a CF of 1. However, a significant improvement in MMLU accuracy over the base model CT is observed with a CF of 2 and 4. The poor accuracy of the token-dropless approach can be explained by a lack of regularization that is introduced implicitly by the CF. For the best trade-off between accuracy and performance, we chose a CF of 4 for our main training configuration. 

\subsection{Choice of Router Algorithm}

\begin{figure}[ht!]
\centering
    \includegraphics[width=0.6\linewidth]{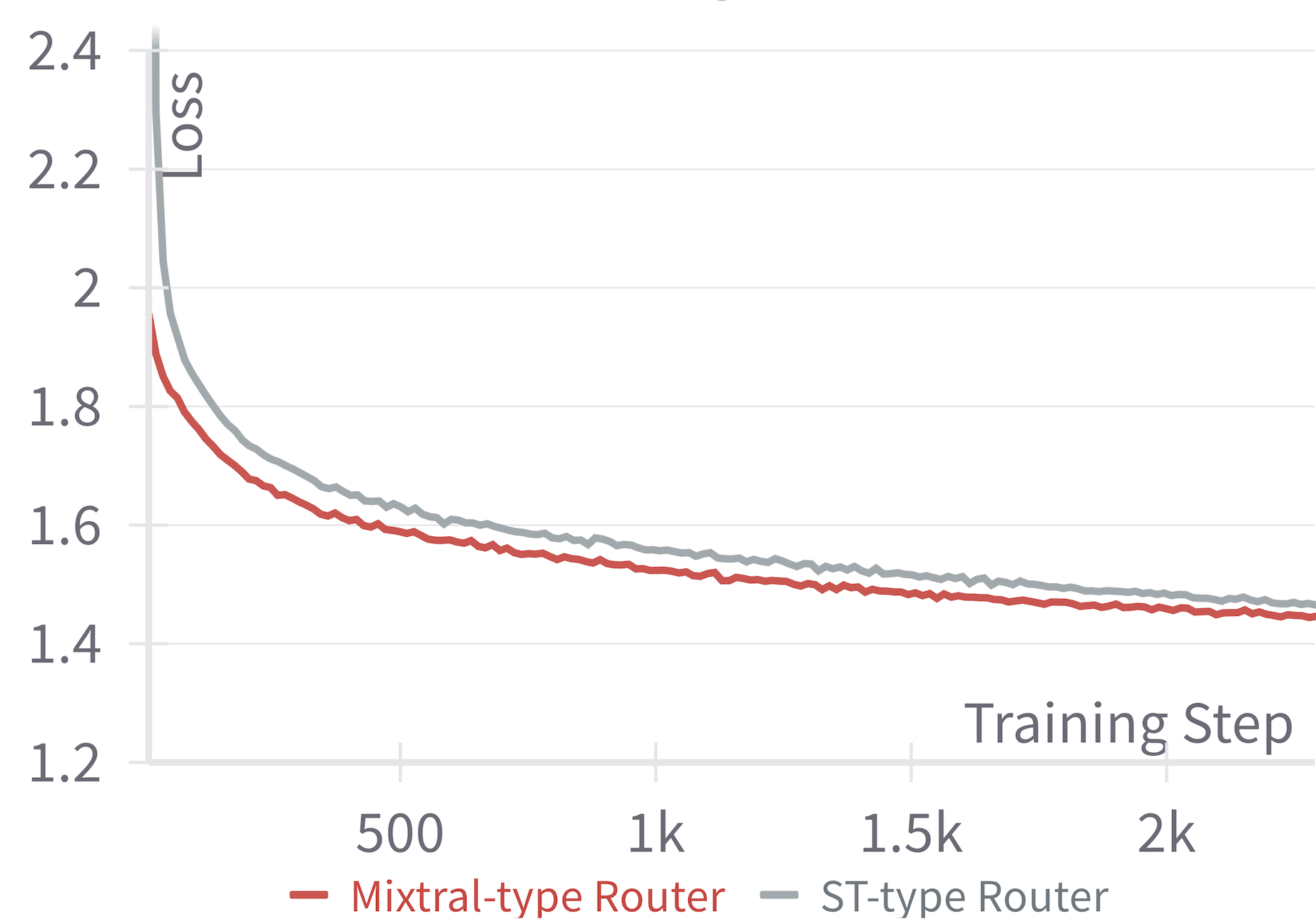}
    \caption{Training loss curve of Mixtral vs ST router types.}\looseness=-1
    \label{fig:router_ablation_loss}
\end{figure}

We compare the effects of the order in which the Softmax and KeepTopK operators are applied within the gating network. The Mixtral~\cite{mixtral} architecture applies the KeepTopK operator first, followed by Softmax, to ensure that the MoE model’s initial forward pass matches the dense model's output. However, this approach sacrifices the information contained in the absolute magnitudes of the router outputs due to the Softmax operation. To address this, the order can be reversed—applying Softmax before KeepTopK, as in~\cite{smoe}—which we refer to as the ST-type router. However, in this reversed configuration, when $1<k<N$, the MoE model’s initial output will no longer match the dense model, potentially leading to training instability and loss of learned representations. This discrepancy may be mitigated with a few training steps, allowing the model to adjust to this change. ~\autoref{fig:router_ablation_loss} compares the training loss curves of Mixtral-type router and the ST-type router. We can see that the Mixtral-type router starts from a comparatively lower loss and converges faster than the ST-type router. Hence we stick to using the Mixtral-type router in our main training configuration.


\section{Conclusion}
In this work, we addressed the challenges and costs associated with scaling large language models (LLMs) by developing an efficient approach to train Mixture-of-Experts (MoE) models. By leveraging pre-trained dense checkpoints to initialize an 8-Expert Top-2 MoE model based on the Llama 3-8B architecture, we demonstrated that high-performing models can be achieved with less than 1\% of the typical pre-training compute. Our experiments validate our choices in capacity factor and routing algorithm, showcasing improved performance on downstream tasks, such as commonsense reasoning and knowledge benchmarks like MMLU. Furthermore, our implementation of online upcycling within the NeMo framework facilitates the effective reuse of pre-trained weights, making MoE model training more accessible to the research community.



\end{document}